# Extending Process Discovery with Model Complexity Optimization and Cyclic States Identification: Application to Healthcare Processes

*Liubov O. Elkhovskaya, Alexander D. Kshenin, Marina A. Balakhontceva, Sergey V. Kovalchuk*

ITMO University, Saint Petersburg, Russia

lelkhovskaya@itmo.ru, adkshenin@itmo.ru, mbalakhontceva@itmo.ru, kovalchuk@itmo.ru

**Abstract.** Within Process mining, discovery techniques had made it possible to construct business process models automatically from event logs. However, results often do not achieve the balance between model complexity and its fitting accuracy, so there is a need for manual model adjusting. The paper presents an approach to process mining providing semi-automatic support to model optimization based on the combined assessment of the model complexity and fitness. To balance between the two ingredients, a model simplification approach is proposed, which essentially abstracts the raw model at the desired granularity. Additionally, we introduce a concept of meta-states, a cycle collapsing in the model, which can potentially simplify the model and interpret it. We aim to demonstrate the capabilities of the technological solution using three datasets from different applications in the healthcare domain. They are remote monitoring process for patients with arterial hypertension and workflows of healthcare workers during the COVID-19 pandemic. A case study also investigates the use of various complexity measures and different ways of solution application providing insights on better practices in improving interpretability and complexity/fitness balance in process models.

**Keywords:** *process mining, process discovery, quality metrics, event aggregation, interpretation, healthcare*

## 1. Introduction

Process Mining (PM) is a newish discipline adopting a data-driven approach and a classical model-based process analysis. It has been actively developing since there is still a demand for better insight into what happens now within an organization. PM is a promising approach to reveal and analyse the real processes existing in all companies today. There are three types of PM: process discovery, conformance checking, and process enhancement [1]. With discovery algorithms, one can automatically obtain a (business) process model from routinely recorded data, an event log. This type of PM is a research topic of most interest [2]. The results of process discovery techniques can be used further in conformance checking and enhancement. A priori process model (discovered from the log or elaborated "by hand") is evaluated on its compliance with data by conformance checking techniques, and its enhancement can be proposed after an analysis of process performance measures. In this study, we address a problem within process discovery. One of the main issues is constructing a model which would be both simple and reflecting actual process behaviour. This often results in the trade-off between quality measures of a model [3]. Ideally, a process model should be understandable and interpretable for both analysts and common users and capture the main way of process execution (if there is no task to find all possible realizations). The problem is most acute when dealing with complex and heterogeneous processes and makes possible to discover a so-called *spaghetti-like* process model [1].

Processes in the healthcare sector are examples of highly varying and distributed processes since they are ad-hoc and healthcare information systems usually are not process-aware [4]. That is why healthcare is the most researched application domain of process discovery techniques [2,5]. Moreover, healthcare organizations need to improve their processes to achieve high-quality care standards in a cost-effective way, and therefore they may benefit from PM solutions. PM community, in turn, needs to elaborate a 'unique value proposition' providing actionable tools which are aware of domain-specific peculiarities and aimed to solve real-world problems [6]. For example, process model structuring often requires domain knowledge. So, an automatic interpretation and a structure analysis of the model are necessary. While there are endeavours and some success in defining and standardizing interpretability in other modelling fields, complex processes with non-trivial domain interpretation are still challenging.

In this paper, we propose an approach for model interpretability based on a meta-states concept and present a technology which extends a PM algorithm with semi-automatic support to model optimization for higher complexity control. We demonstrate our solution applicability within the healthcare domain, where processes are best suited for model comprehension enhancing and from which the idea of the concept is originated. Despite the concrete study case, we believe the approach is adaptable to other domains, and it is broadly considered as an extension of a process discovery technique.

## 2. Related works

### 2.1. What are Complexity and Interpretability in PM?

In different system modeling domains, research and development are mainly aimed at high accuracy of model fit, i.e., capturing dependencies in data, while *interpretation* is receiving a little attention. It is believed interpretation can increase the trust to the models, which is particularly acute for black-box models, even though it may affect predictive accuracy. This evoked a new line in machine learning [7] and artificial intelligence [8], which models can already learn complicated data patterns and make precise forecasts. Yet, however, there is no strict definition of interpretability. When it comes to a system insight, one can also reason about its *complexity*. Like interpretability, complexity is context-sensitive and can be hardly defined universally. There are many definitions and meanings of complex systems and complexity measures, where the last can be derived into two major

classes — computational and system-related [9]. Interpretability and complexity are closely related: a more straightforward description produced by the system tends to be better construed.

Comprehensive ability of a human of a business process model can be influenced by many factors ranging from personal characteristics to elements of Gestalt theory [10]. They are chosen process modelling notation and its features (visual expressiveness, semantic transparency, etc.), the size of the model (number of elements, diameter), modularity and structuredness (use of constructions with split/join usually improves model understanding), decomposition (ways to hide unimportant information from the user, thereby improving the quality of the model), etc. Within PM studies, Mendling et al. [11] investigated aspects that may influence on process model comprehending. The authors used questionnaires with several process models that were filled by students of three universities. The students were taking or completing classes on PM and had different levels of subject knowledge. The study revealed that in addition to personal factors, model size is correlated with its comprehending. A similar result was obtained in another empirical study [12]: larger distance of a process structure tree, more challenging process behaviour to perceive. Statistical analysis also indicated the ranking of cognitive difficulty when understanding different relationships between model elements. In [13], process model complexity was defined as the degree to which a process is difficult to analyze, understand, or explain. The model structure, therefore, can be the strong evidence of its complexity. In PM, the complexity term is often associated with logical blocks such as AND-, XOR-, OR-splits, and loops presented in the model. These split constructs can be evaluated and quantified using different measures, e.g., control-flow complexity [13], entropy-based uncertainty [14], and others [15]. Their sum is the overall architectural complexity or uncertainty. Other measures of complexity can be derived if one will consider process models from the prospectives of neighboring disciplines [16]. In this study, we look at a process model as a graph structure and exploit several network complexity measures to find an understandable and interpretable model.

## 2.2. Towards Process Model Optimization

Addressing the problem of optimal process model, we mean constructing a model, which is simple for understanding and accurately captures process behaviours at the same time. Ideally, finding a balance between these quality components should be as automated as possible, and simplicity of the model can be achieved through its elements' abstraction/aggregation. Table 1 contains studies which affect the issues to a certain extent. Most of the papers are for the last 6 years, and there are several earlier works as baselines. The proposed solutions are classified on the type of approach used and evaluated on the degree of its applicability to the aspect concerned.

Log pre-processing may be a good starting point for enhancing both process visualization and model precision. This category of techniques for model enhancing is applied before the very discovery retrieving log information directly. *Suriadi et. al* [17] demonstrate a systematic approach to event log preparation based on imperfection patterns of records. Common quality issues compose such patterns. They are form-based event capture, unanchored event, inadvertent time travel, etc. The authors describe each pattern and show real-life cases, and then propose possible solution and its side-effects. For example, distorted labels, the most frequently observed pattern according to the questionnaire results, negatively impact the readability and validity of process mining results. The activities which have the same semantics but do not match each other due to incorrect data entry or ununified recording systems should be treated as one. This could be done by agreed to letters capitalization, similarity string search, manual interventions (e.g., using a knowledge base [17], ontology, or rule base [18]). More intelligent techniques such as topic modelling [19] or conditional random fields [20] can facilitate moving from low-level to high-level events aiding comprehensibility of discovered process models. In [21], the authors propose an interval-based selection method to filter outliers, which are repeated events within the specified time period. Event outliers are defined based on the distribution of time intervals of consecutive events of the same activity. The time perspective regard is potential for log pre-processing step; the proposed method has improved model precision without reducing its fitness. The opposite option to mine repeated activities based on the contextual information is proposed in [22]. The reason for mining duplicate tasks is enhancing model understandability and clarity, e.g., by its unbranching. These approaches only facilitate the discovery of a better-quality model but do not guarantee its optimization.

The second type of approaches simplify process models during or after their discovery by behaviour filtering. These approaches are mostly implemented with manual methods. It is common to allow the user inspecting some threshold-fixed model first, and manual adjusting its parameters then. The Fuzzy [23] and Skip [24] miners are prime examples of the discovery algorithms adopting such strategy. This technique is commonly used in the discovery algorithms which output is a directly-follows model to reduce its complexity through abstraction and aggregation [25]. Fodina [22] and BPMN Miner [26] also incorporate dependency information like in Heuristics Miner [27] to filter infrequent activities and arcs and to deal with split/join constructions and binary conflicts. Additionally, the authors [26] utilize both originator and control-flow perspectives in the discovery to group activities into swimlanes, which can be collapsed for more abstraction. Self-loops and short loops can cause a problem while analysing concurrency relations between tasks. It can affect not only correctness of the discovery but model complexity. Approaches which deal with such structures are proposed in [28,29]. The proposed algorithms are able to produce simple models while balancing fitness and precision. In [30], the authors addressed the issue of mixed-paradigm process mining, which aims to discover a precise and comprehendible process model at the same time. They introduced Fusion Miner and proposed a new metric called activity entropy. It captures activities connected with most of the other activities, i.e., identifies weak dependencies. This way, considering behavior types that an event log contains, one can change an input parameter of the miner to achieve the balance between procedural and declarative constructs in a model. The described above approaches address both optimization and abstraction issues of process discovery but with manual methods. They usually require many parameters configuration with several trials to adjust a model at the desired level of granularity.

Since the number of model elements influences its comprehending, reducing the variety of events is necessary. For example, collateral events, which are multiple events referring to one process step, may noise the data, and they need to be aggregated [17]. The time perspective should be regarded to merge activities occurred together within a specified time period. The same is for events captured by the form and affected by the same timestamps, such as a set of patient blood tests. The relation of two subsequent events can be measured by correlation metrics. Correlation may be determined with respect to timeframe within which events occur one after another, originator or activity names similarity, and data perspective. The approach proposed in [23] groups highly correlated and less-significant nodes into clusters, therefore, abstracting them into one logical or high-level activity. A discovery technique proposed in [31] captures data hierarchies, which presents in software systems. Within these systems, logged execution calls or code architecture itself can be utilized in the designed hierarchy. The hierarchy can be reflected in the model at different levels of depth, i.e., hiding or unfolding its parts interactively by the user, therefore, simplifying or complicating the discovered model. The technique with different heuristics is evaluated on the examples of software event logs; it has overall positive impact on the model quality. The hierarchy also can be seen in healthcare events represented as ICD-10 codes [32]. Aggregation as a post-processing step is demonstrated in [33]. Here, the authors merge similar behaviour after an alternative choice using the folding equivalence. This typically generalizes the behaviour of the net and reduces its complexity. So, aggregation techniques are good options to abstract a process model and possibly simplify it at the costs of losing some information.

Unsupervised learning has found uses in different tasks and fields including PM. It is common trace clustering to be applied in a discovery step. In this regard, the event log is divided into sub-logs to produce more accurate process models for each cluster. However, the clustering techniques need a robust similarity metric [34]. They also should incorporate PM quality measures since data mining criteria only assume the discovery results improvement. Thus, the authors [35] propose a semi-supervised method with a selection step for a greedy accuracy optimization for each cluster. Despite the high computational complexity, the novel approach has performed better than MRA-based clustering in terms of accuracy. A "slice-mine-dice" technique is proposed in [36] to cluster traces reaching the specified complexity threshold. The results showed the improvements in the number of clusters and process model cumulative size in comparison to three existing trace clustering methods. Contextual information is utilized in [37] in a modified k-medoids clustering algorithm. Standard process is assumed to be more frequent and stable in a time perspective. So, occurrence frequencies and overall cycle times are included to support grouping of similar process variants. Using the first $k$ most frequent processes as medoids has improved the clustering quality in several heterogeneous cases. However, identifying the proper value for $k$ is still an open issue. In [38], the coefficient of variation is used to determine the number of clusters of clinical pathways represented as state (department in a hospital) sequences. The clusters with insufficient number of sequences were discarded before the discovery, where identification of typical pathways is based on an alignment algorithm from bioinformatics. Clustering can be performed not only for traces but for events too. A two-level clustering approach is proposed in [39] to categorize complex events first, and to group obtained processes then, also to distil clinical pathways in real cases. Such approaches aid abstraction of undesired details and, therefore, simplify models but mostly in uncontrolled manner.

Automated control of finding a simple and/or precise process model can be explicitly defined as an optimization problem. For example, Camargo et al. [40] optimized the accuracy of a business process simulation model discovered by the Split Miner. They searched for optimal hyper-parameters of the miner to maximize the accuracy measured using a timed string-edit distance between the original and simulated by the model event logs. However, there is always the trade-off between the several quality dimensions [3], where the most common one is between the model complexity and its fitness/precision. Within an integer linear programming, the authors [32] define process model optimization problem mathematically, where a linear constraint is a complexity threshold (a maximum number of nodes and arcs) and a replayability score aimed to be maximized. They solve the optimization task with a tabu search algorithm utilizing elements' frequencies to quickly identify promising moves. The approach outperforms commercial tool's results in terms of replayability for small and middle complexity (from 1 to 50 nodes). A similar problem is formulated in [41] but with including time-related information in the replayability score of discovered grid process models. The tabu search with optimized edges is showed to be more efficient than other methods for a small event diversity. In [42], rule discovery hybrid particle swarm optimization is proposed to find near-optimal process models. Here, the simulated annealing is applied on each particle position when it is updated. Additionally, the authors apply rule discovery to get the top particles which meet the criteria and, therefore, formulate an optimization problem. The proposed method has the best results in terms of average fitness and number of iterations in comparison with a classical particle swarm optimization method and hybrid one. It also has the potential to get the higher comprehensibility performance due to a rule discovery task. A multi-objective optimization via Pareto optimality is addressed in [43]. The reason of a Pareto front is that the quality dimensions are mutually non-dominating, and the user can choose the desired trade-off visually (for 3 dimensions and less). In [44], the authors propose a modified genetic algorithm for process discovery, called ProDiGen. It deploys a hierarchical fitness function to evaluate individuals in a population. The fitness function incorporates both completeness, precision, and simplicity of a mined model. ProDiGen correctly mined the original models in most cases; the obtained models are simple, complete and precise. It has a better performance than other PM algorithms and comparable to Genetic miner's [45] computational times for the balanced and unbalanced logs with different workflow patterns and levels of noise. We believe, therefore, the clearly formulated objectives as an optimization task are the best option to automate and control the discovery step.

Table 1 – Studies addressing process model optimization problem through different approaches

| Approach | Studies | Optimization | Methods |
| --- | --- | --- | --- |
| Log pre-processing | *Suriadi et al.* [17], *Leonardi et al.* [18], *Chiudinelli et al.* [19], *Tax et al.* [20], *Alharbi et al.* [21], *Broucke et al.* [22] | ± | Outliers (events, traces) detection and removal, region-based methods for repeated tasks, topic modelling, sequence labelling |
| Behaviour filtering | *Günther et al.* [23], *Batista et al.* [24], *Weerdt et al.* [26], *Broucke et al.* [22], *Leemans et al.* [25,31], *Augusto et al.* [28], *Sun et al.* [29], *De Smedt et al.* [30] | ✓ (manual) | Activity, precedence relation, cycles, and split/join filtering; conflict resolution; attribute accounting |
| Aggregation | *Suriadi et al.* [17], *Günther et al.* [23], *Leemans et al.* [31], *Prodel et al.* [32], *Fahland et al.* [33] | ± | Hierarchical event structure (e.g., ICD-10 codes, software code architecture), correlation metrics, model construction folding |
| Clustering | *Delias et al.* [34], *Weerdt et al.* [35], *García-Bañuelos et al.* [36], *Becker et al.* [37], *Funkner et al.* [38], *Najjar et al.* [39] | ± | Trace and event clustering |
| Optimization problem | *Prodel et al.* [32,46], *Camargo et al.* [40], *De Oliveira et al.* [41], *Effendi et al.* [42], *Buijs et al.* [43], *Vázquez-Barreiros et al.* [44] | ✓ | Linear programming, Pareto optimality, particle swarm optimization, etc. |

## 3. Conceptual approach

### 3.1. Basic idea

The proposed approach is based on the extension of PM techniques with several procedures, which are illustrated in fig. 1 and discussed in detail furtherly in this section. Process discovery is followed by complexity and fitness estimation procedures ("1" in figure 1). The complexity level is estimated using structural and representative characteristics of the identified process model. The value of fitness is estimated through a comparison of event log coverage with a particular process model.

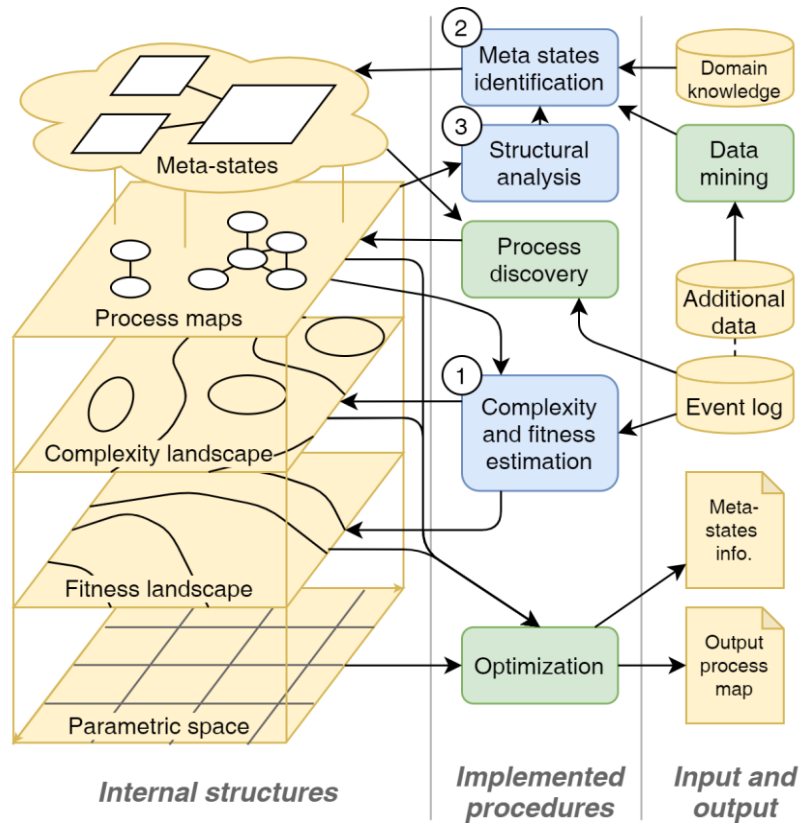

Figure 1 – General approach for PM procedures extension

The estimated values are furtherly used to perform optimization within a parametric space for a process identification algorithm. In the general case, the optimization problem here is a multi-objective problem where objectives in most cases contradict each other, and the higher complexity corresponds to better fitness.

To introduce higher-level states and domain interpretation, we propose the procedures for meta-state identification. The meta-states identification ("2" in figure 1) has various sources of knowledge available for use during the identification problem. The primary source is a structural analysis of process maps ("3" in figure 1) with the identification of cyclic states, which are often collapsable to a single meta-state. Second, we can exploit the data available in event logs (event attributes, case parameters, etc.) to identify states of the process. Usually, the data can be analyzed using data mining techniques. Finally, domain knowledge can also be applied to introduce information on states specific to a particular process or system where the process evolve.

The meta-states identification could be excessively used for analysis and interpretation of a process model. However, within the presented work, we are focused on the explicit introduction of meta-states into the process model to bring more expressive power with the lower model complexity.

### 3.2. Process discovery algorithm

In this subsection, we introduce an extended algorithm for process model discovery. However, it is necessary to provide basic definitions and a general view of the problem where do we start first. Every process-aware information system that records run-time behavior has an *event log*. An event log is a file that contains information about process execution. Each record is an *event* with associated data: *timestamp* of its start and completion, an *activity* and *resource* that executes this activity, and a process *case id* (instance) the record belongs to. They are the minimal items for compiling a log. However, if activities are considered to be atomic, i.e., have no duration, the last item is needed only for defining the order of them and can be skipped if we *a priori* know data is stored according to a timeline. We group an ordered set of events containing only activity names into cases, that represent single process runs. This "flat" event log is used as an input for process mining in our discovery algorithm.

While an event log is an input, the algorithm's output is a (*business*) *process model*, or a *process map*. In our case, a process model represents a formal graphical description of the actual process flow, i.e., the precedence of events, where nodes are activities and edges are ordered relationships between them. Below (fig. 2), we provide a general scheme of the solution for obtaining such a result, where dashed arrows are optional discovery steps, and main parameters are listed in callouts.

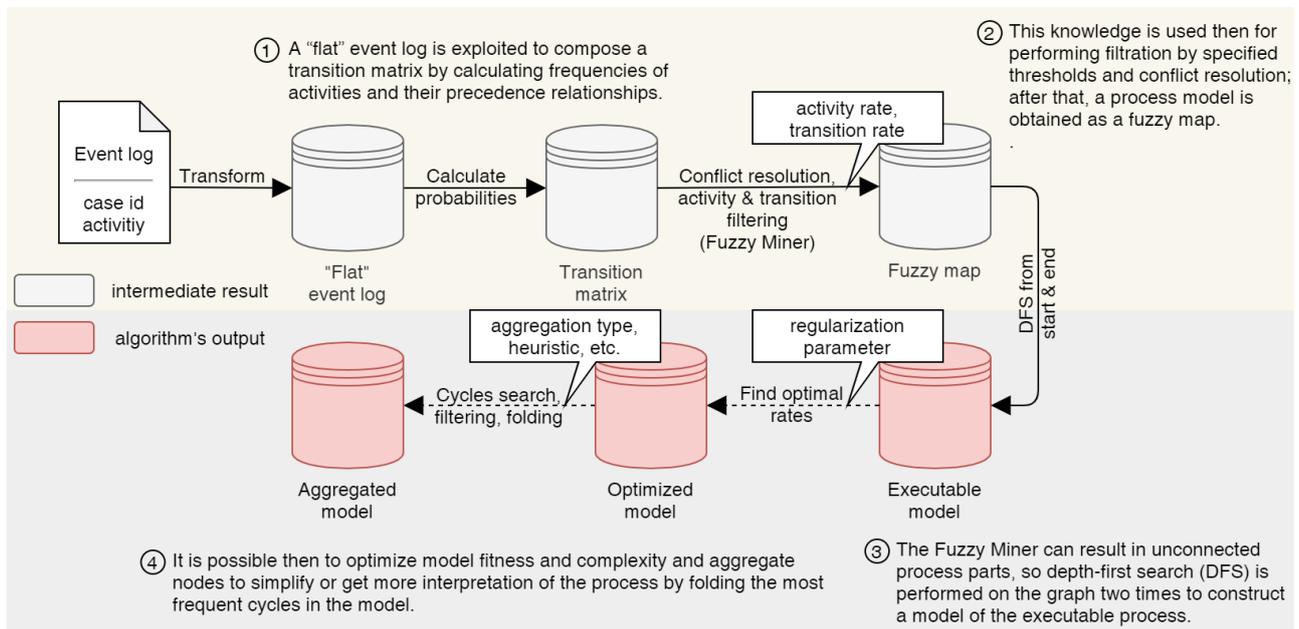

Figure 2 – General scheme of the algorithm's workflow

The next section presents the precise details of the proposed algorithm implementation. Steps 1-3 are the basics of the algorithm's workflow and described in following subsection 4.1. Within step 4, we define an optimization problem in subsection 4.2 and propose an approach to model abstraction by folding significant cycles in subsection 4.3.

## 4. Implementation of the extended algorithm

### 4.1. Model discovery

The proposed algorithm for discovering process models includes the basics of the Fuzzy Miner [23], which, in turn, comes from one of the first discovery techniques Markov. The main idea is to use the theory of Markov discrete random processes to find the most probable transitions between events. The fundamental metric is a *significance* that can be determined for event classes (i.e., activities) and binary precedence relations over them (i.e., transitions). Significance is the absolute or case frequency of activities or transitions that are occurred in the event flow. It measures the relative importance of the behavior, i.e., events or precedence relations that are observed more frequently are deemed more significant. We use case frequency in conflict resolution, when two events may follow each other in any order in the event log, and process simplification, i.e., activity and transition filtering, and absolute frequency is for statistics visualization.

The algorithm constructs a directly-follows graph (DFG) like a finite state automaton but with activities represented in nodes rather than in edges, i.e., Fuzzy map. This representation is more intuitive and understandable and can be easily transformed into other notations. Within the used visual notation, the green node ("start") indicates the beginning of the process and shows the total number of cases presenting in the log, and the red node ("end") is related to a terminal state. The graph's internal nodes and edges show the absolute frequencies of events and transitions, respectively: more value, darker or thicker element.

However, fuzzy logic does not guarantee a reachable graph (see example in Fig. 3) which is desired to see the complete behaviors of process traces. So, we modify model construction by performing the depth-first search (DFS) to check whether each node of the DFG is a descendant of the initial state ("start") and a predecessor of the terminal state ("end"). If the model does not match these conditions, we add edges with respect to their significance to the model until we get a reachable graph. This way, we overcome the possibility of discovering an unsound model (without the option to complete the process). Despite other DFG limitations [47], it permits cycles, which are crucial in our concept of meta-states, and it is suitable for unstructured and complex processes, which exist in healthcare, due to constructing a model at different levels of details.

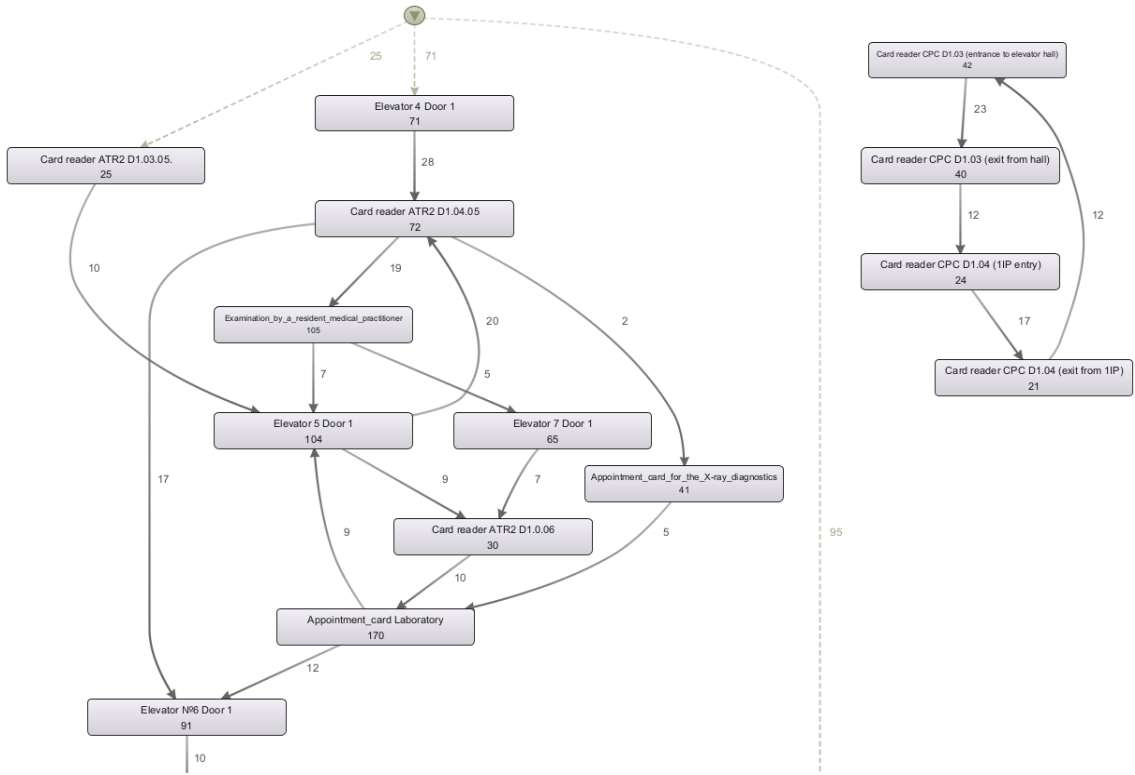

Figure 3 – Example of disconnected model obtained by Disco[1].

We show an example of a process model obtained by the proposed algorithm, adjusted manually, and resulted in 100% and 5% of activity and transition rates, respectively, in figure 4. It means that only activities and transitions with significance more than or equal to 0.0 and 0.95, respectively, are included in the model. In other words, we aim to see only the main paths with all event variations. We will further attribute rates to the model as $r_a/r_t$ where $r_a$ is an activity rate and $r_t$ is a transition rate. The process model was discovered from the event data of a remote monitoring program for patients suffering from hypertension. More details about this and other datasets and process models for these event logs are given in Section 6.

---

[1] fluxicon.com/disco/

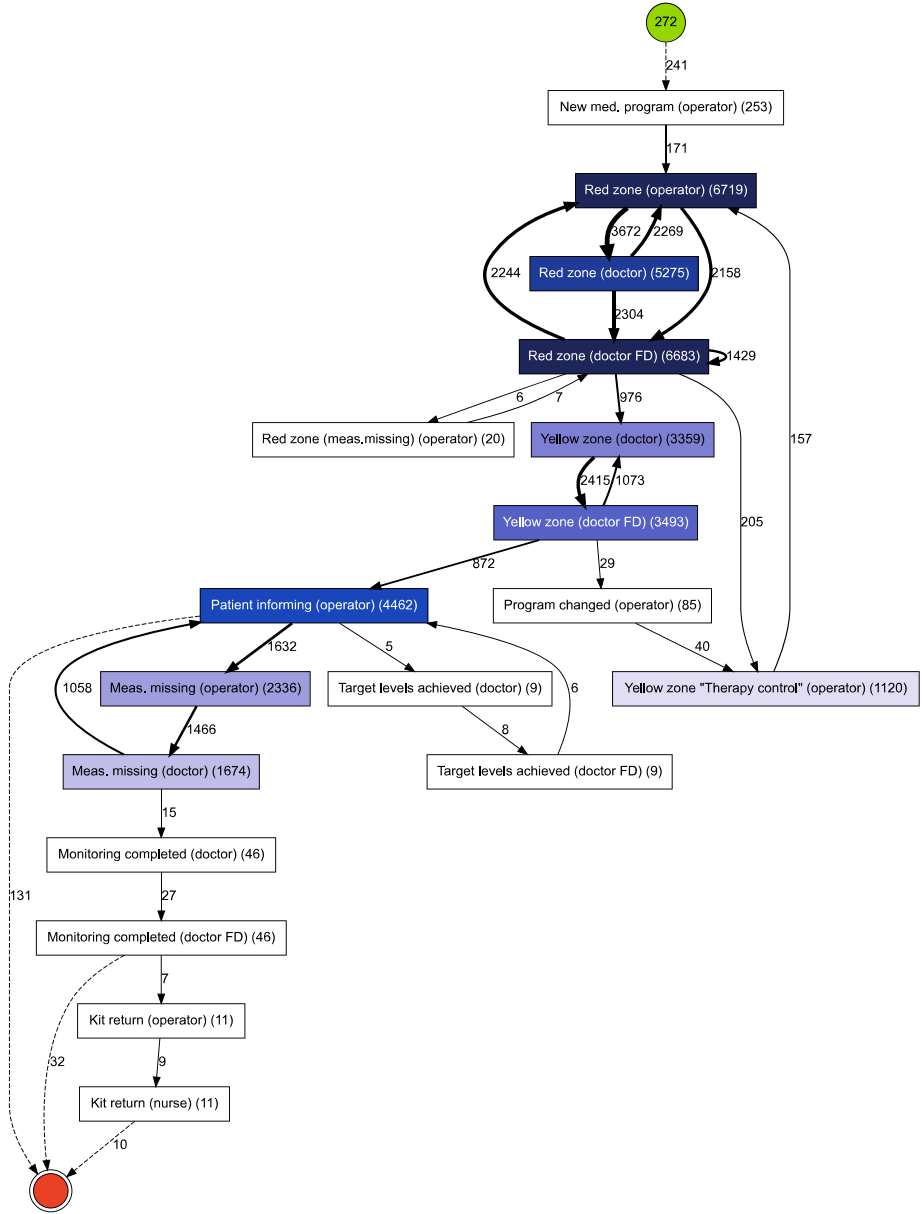

Figure 4 – Model of the remote monitoring process performed for the patients with hypertension

### 4.2. Model optimization

In primary usage of the algorithm, one can change process model detail by tuning activity and transition rates and move from the simplest model to complex and fullness one. However, it is preferable to automatically obtain a model when it comes to the tool's adaptiveness and massive processing. Therefore, we attempted to apply machine learning formalization in PM and defined the problem of automatically discovering an optimal process model.

Let $p$ be the algorithm for discovering a process model from an event log. The set of traces $L = \bigcup_{i=1}^{N} \sigma_i$ is an event log, where $\sigma_i = \langle e_1, e_2, \ldots, e_{k_i} \rangle$ is the $i$-th process execution instance (trace) of the length $k_i$ and $\sigma_i(j) = e_j$ is an event from the set of events $X$ which the log can contain, and the log size, therefore, is $l = \sum_{i=1}^{N} k_i$. An event can be defined as a set of attributes (activity type, resource type, time stamp, etc.) but here and further we imply an atomic event log, i.e. each event is an activity. We also assume that activities within a trace are ordered by the time registering in a system. Let $p(L, \bar{\theta}) = M = \langle V, E, v_{start}, v_{end}, sig \rangle$ be a process model, which is a DFG, discovered by the algorithm $p$ with parameters $\bar{\theta} = (\theta_1, \theta_2)$, activity and transition rates, respectively, from the event log $L$, where:

- $V \subseteq X$ is a set of nodes, $|V| = n$;
- $E \subseteq V \times V$ is a set of edges, $|E| = m$;
- $v_{start}$ is a "start" (initial) node;
- $v_{end}$ is an "end" (terminal) node;
- $sig: X \cup (X \times X) \to (0, 1]$ is an activity and transition significance defined as a case frequency, a fraction of traces that contain an activity or transition:

$$sig(x) = \frac{\sum_{j=1}^{k} 1_{\sigma_j}(x)}{k} \tag{1}$$

for an element $x$, where $1_\sigma(x)$ is an indicator function that equals one if an element $x$ is contained in a trace $\sigma$ and equals zero otherwise. For an activity, it is defined as follows:

$$1_\sigma(x) = \begin{cases} 1, \exists i = \overline{1, n}: x = \sigma(i), \\ 0, \text{otherwise}; \end{cases} \tag{2}$$

and in case of transition:

$$1_\sigma(x) = \begin{cases} 1, \exists i = \overline{1, n-1}: x = \langle \sigma(i), \sigma(i+1) \rangle, \\ 0, \text{otherwise}; \end{cases} \tag{3}$$

Let $P = \{p(L, \bar{\theta}) | \bar{\theta} \in \Theta\}$ be the process model space, where $\Theta$ is a domain of the algorithm parameters. Here we consider $\bar{\theta} = \langle r_a, r_t \rangle$. Thus, $\Theta = [0; 100] \times [0; 100]$. One needs to find an algorithm $p \in P$ (more precisely its parameters) that maximizes $Q$ on $L$:

$$Q(p, L) = (1 - \lambda) \cdot F + \lambda \cdot (1 - C_\mathcal{J}) \to \max_{\bar{\theta}}, \tag{4}$$

where

$$F = \frac{1}{|L|} \sum_{\sigma \in L} \left( \frac{1}{|\sigma|} \sum_{i=1}^{|\sigma|} z_{i,n} - \alpha \delta(\sigma, s^*) - \beta \frac{\phi(M, \sigma, s^*)}{n} \right)^+, \tag{5}$$

$z_{i,n}$ is a binary variable equal to 1, if event $i$ is represented by node $n$, $\phi$ is the number of *forced transitions*, $\delta$ is *event skipping indicator*, $s^*$ is the subtrace of all events represented by the process model $M$ [32],

$$C_\mathcal{J} = \frac{\mathcal{J}(p(L, \bar{\theta}))}{\mathcal{J}(p(L, \bar{\theta}^{100}))}, \tag{6}$$

$$\mathcal{J}(p(L, \bar{\theta})) = \frac{m}{n}, \tag{7}$$

and $E_0: p(L, \bar{\theta}^0) = \langle V_0, E_0, v_{start}, v_{end}, sig \rangle$, $\bar{\theta}^0 = (0, 0)$; $\bar{\theta}^{100} = (100, 100)$.

In this optimization problem (4), an objective function includes fitness (5) and complexity (6) terms where $\lambda$ is the regularization parameter to weight them. Thus, one can discover a process model optimized in one of these senses or both.

The representativeness of a process model regarding a log is measured by the replayability [32,41], also called fitness or fidelity. This metric works fine with flexible logs with highly diverse and complex behaviors of the traces. It also overcomes DFG issues with alignments. Replayability is directly related with model complexity (6)-(7): a model with higher complexity allows for more traces and therefore higher replayability, making these measures contradictory objectives. A remarkable feature of replayability (5) is that it is scaled to be a number in [0, 1]. So, it can be combined with the scaled complexity term (6) in one objective function (4). Here and after we assume $\alpha = 0.5N^{-1}$ and $\beta = N^{-1}$, where $N$ is defined below as the number of unique activities in the log.

Complexity function could be performed as one of the network complexity measures [48–50]. In this study, we compare different measures, one of which is the Shannon entropy. Entropy can be measured across various network invariants [49]. For the current study, we chose a flattened adjacency matrix as a random variable $X$ with two possible outcomes (0 or 1) to measure the Shannon entropy $H(X)$:

$$\mathcal{J}: H(X) = -\sum_{i=1}^{n} p(x_i) \log_a p(x_i). \tag{8}$$

Alternatively, we have introduced several structural complexity measures as follows:

$$\mathcal{J}: K_n = \frac{m}{n(n-1)}, \tag{9}$$

$$\mathcal{J}: R = \frac{1}{2} \cdot \left(\frac{m}{M} + \frac{n}{N}\right), \tag{10}$$

where $N$ and $M$ are the numbers of unique activities and transitions in the log, and $n$ and $m$ are the numbers of the activities and transitions (unique) presented in the model, respectively. Measures above, thereby, explain how many elements were displayed in the model among theoretical or possible ones. They also can specify model complexity. However, it may penalize model a lot and shorter process behavior. So, we currently chose a simple graph measure, an average degree. In a directed graph, it is just the number of edges divided by the number of nodes:

$$\mathcal{J}: AD = \frac{m}{n}, \tag{11}$$

This measure is well suited to our aims: we want to reduce the number of transitions and persist only significant ones while lengthening paths through the model by remaining more activities. This way, one may achieve more transparent, not confusing as well as a consistent process model. In Section 5, we present an experiment on complexity optimization with different measures and show examples of automatically discovered in such way process models.

### 4.3. Discovering meta-states

In this subsection, we introduced an approach for process model abstraction and simplification. Simplification of the process model can be done not only by node and edge filtering but also by events aggregation. In [23], the authors proposed such model abstraction iteratively aggregating highly correlated (in context sense) but less-significant nodes. However, in some fields, e.g., in healthcare, it makes sense to propose another method of abstraction. It is very likely cycles present in the model, and this can signify distinct process parts from the perspective of *for whom* the process is performed. In healthcare, the cycles may represent routine complex of procedures or repeated medical events for patients, i.e., objective being in some stage of process execution or a *meta-state*. We assume a simple cycle to be a meta-state if the probability of its occurrence in the log exceeds the specified threshold, i.e., a cycle significance, as in case of activities and their correspondence relations filtration. One can get new knowledge about the process execution by distinguishing the most significant cyclic behavior and exceptions. We clarify how meta-states are identified in an event log via pseudocode below.

---
Algorithm 1. Searching cycles and counting their frequencies in an event log
---

**procedure** CyclesSearch(Log)
Input: "Flat" event log Log composed of process cases
Output: Set of (simple) cycles $cycles$ found in event log Log, Absolute abs[$c$] and case cse[$c$] frequencies of each cycle $c \in cycles$

cycles ← [ ]
k ← 0
**for** all cases $t \in$ Log **do**
  **for** all unique activities $n \in t$ **do**
    case_cycles ← [ ]
    $i$ ← 0
    $j$ ← 0
    **while** $i <$ length of $t$ **do**
      **if** $t[i] = n$ **then**
        case_cycles[j] ← $i$  // Positions of activity $n$ in case t
        j ← $j + 1$
      end **if**
      $i$ ← $i + 1$
    end **while**

    $i$ ← 0
    **while** $i + 1 <$ length of case_cycles **do**
      c ← $t$[from case_cycles[$i$] to case_cycles[$i + 1$] $-$ 1]  // Part of case $t$ that starts and
                                                                               ends with activity $n$
      **if** length of $c$ = number of unique activities $a \in c$ **then**
        **if** $c \notin$ Cycles **then**
          cycles[k] ← $c$
          abs[$c$] ← 0
          cse[$c$] ← 0
          $k$ ← $k + 1$
        end **if**
        abs[$c$] ← abs[$c$] + 1
        cse[$c$] ← cse[$c$] + 1 // if $c$ was not found earlier within a case $t$
      end **if**
    end **while**

```
      end for
   end for
end procedure
```

---

Algorithm 2. Identification of significant cycles (meta-states) in an event log

```
procedure FindStates(Cycles, CycleFrequencies, NumberOfCases, MetaStateSignificance)
Input:  Set of (simple) cycles Cycles found in a process model by DFS;
        Case cse[c] ∈ CycleFrequencies frequency of each cycle c ∈ Cycles;
        Number of cases NumberOfCases in an event log;
        Required significance of cycle MetaStateSignificance to be defined as meta-
        state
Output: Set of meta-states (significant cycles) MetaStates

for all cycles c ∈ Cycles do
   if length of c > 1 then
      if cse[c]/NumberOfCases ≥ MetaStateSignificance then
         MetaStates ← ADD(c)
      end if
   end if
end for
end procedure
```

---

We propose two types of aggregation. In the first type, nodes included in meta-states are allowed to be present distinctly in a process model. We can call this aggregation as "outer". In contrast, "inner" aggregation redirects all relationships of single events to corresponding meta-states. Here, we got different ways how to redirect relations: to all meta-states that contain such an event or to most frequent one.

According to notations proposed in the previous subsection, we give a formal description of model aggregation. Let $M = \langle V, E, v_{start}, v_{end}, sig \rangle$ and $M' = \langle V', E', v_{start}, v_{end}, sig' \rangle$ be a process model before and after significant cycles folding, respectively, and let us introduce the following notations:

- $\tilde{V} = \{\tilde{v} = \langle \tilde{v}_1, \tilde{v}_2 \dots \tilde{v}_k \rangle | \langle \tilde{v}_j, \tilde{v}_{j+1} \rangle \in E \; \forall j = \overline{1, k-1}, \forall k \leq n, \langle \tilde{v}_k, \tilde{v}_1 \rangle \in E\}$ – meta-states, i.e., significant cycles found in the process model $M$, and $\tilde{v}(i) = \tilde{v}_i$,
- $V^+ = \{v \in V | \exists i \; v = \tilde{v}_i, \tilde{v} \in \tilde{V}\}$ is a set of meta-state vertices,
- $V^- = V \setminus V^+$ is a set of vertices not appeared in meta-states,
- $\tilde{E} \subseteq E \cup (V \times \tilde{V}) \cup (\tilde{V} \times V)$ is a set of edges obtained for the event log with collapsed cycles.

Then $V' \subseteq V \cup \tilde{V}$, $E' \subseteq \tilde{E}$ for outer aggregation, and $V' \subseteq V^- \cup \tilde{V}$, $E' \subseteq (V^- \times \tilde{V}) \cup (\tilde{V} \times V^-) \subseteq \tilde{E}$ after inner joining with updating significance for redirected edges as follows:

$$sig'(\langle u, v \rangle) = \frac{\sum_{j=1}^{k} \vee_{q:q=v(i)} \left( r_*(q,v) \wedge 1_{\sigma_j}(\langle u,q \rangle) \right)}{k} \; \forall u \in V^-, v \in \tilde{V}, \qquad (12)$$

$$sig'(\langle u, v \rangle) = \frac{\sum_{j=1}^{k} \vee_{q:q=u(i)} \left( r_*(q,u) \wedge 1_{\sigma_j}(\langle q,v \rangle) \right)}{k} \; \forall u \in \tilde{V}, v \in V^-, \qquad (13)$$

where $r_*$ is defined for filtering aggregation function that is defined in two forms:

$$r_{all}(v, \tilde{v}) = \begin{cases} 1, \exists i: v = \tilde{v}(i), \\ 0, \text{otherwise,} \end{cases} \forall v \in V^+, \tilde{v} \in \tilde{V}, \qquad (14)$$

$$r_{freq}(v, \tilde{v}) = \begin{cases} 1, \underset{v' \in \tilde{V}}{\text{argmax}}\ sig(v') = \tilde{v}, \\ \quad \exists i: v = v'(i) \\ 0, \text{otherwise,} \end{cases} \forall v \in V^+, \tilde{v} \in \tilde{V}. \qquad (15)$$

Formulas (12) and (13) depict how we "hide" transitions between events, one of which meta-states absorb, i.e., recalculate their frequencies according to $r_*$ (14)-(15). After meta-states discovery, the event log is rebuilt, as in example illustrated in figure 5. The activities included in meta-states are not considered when we mine a process from the event log in the case of inner joining. Their precedence relations are redirected to corresponding meta-states determined by $r_*$. We show an example of how the proposed technique can transform a model in figure 6: (a) the initial process model has two simple cycles (BC and BCD); (b) if we assume they appeared in the event log more than in half of the cases, they are significant and may appear in the model as nodes along with the activities which compose these cycles in the case of outer aggregation; (c) performing inner joining with $r_{all}$ hides activity C, which is an element of the significant cycles, and incorporates frequencies of transitions associated with C (A→C) with frequencies of transitions to or from all meta-states containing C (A→BC, A→BCD); (d) inner joining with $r_{freq}$ is similar to the previous case but recounts frequencies of only most significant meta-states (e.g., BCD).

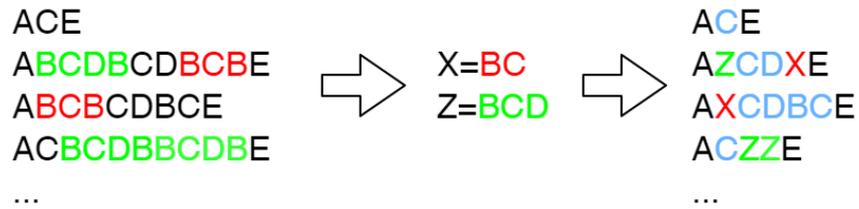

Figure 5 – Cycles collapsing in event log. Activities colored with the blue present in meta-states but do not compose them in the log: they will not be included in the model in case of inner aggregation.

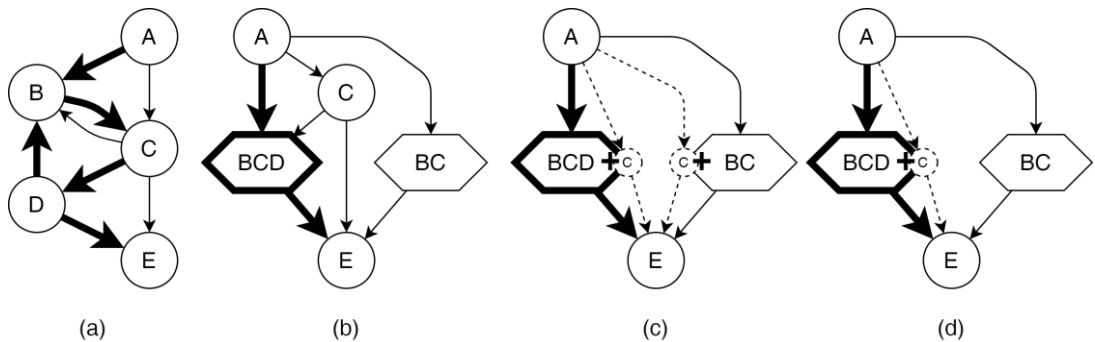

Figure 6 – Possible rebuilding of a process map with cycles (a) initial map; (b) outer joining; (c) inner joining with $r_{all}$; (d) inner joining with $r_{freq}$

### 4.4. Software implementation

To implement the proposed algorithms, we have developed a python library ProFIT[2] (*Process Flow Investigation Tool*) for process mining with a higher degree of automation in complexity control. The library is considered as extendable software solution which can be applied in various contexts and problem domains. We implemented "Observer" OOP pattern in the main class *ProcessMap*, where "observers" are *TransitionMatrix*, *Graph*, *Renderer* that are updated when data or parameters were changed. These three classes store formal information about process structure, i.e., appropriate order of event relations and transition probabilities, sets of nodes and edges in a graph, set of elements, and their arrangement in 2-D space. With knowledge discovered from an event log by a single method of *TransitionMatrix*, we perform process mining in *Graph*, where the main algorithm and approaches are employed. *Renderer* object transforms an obtained model from graph notation into DOT language and then visualizes it by Graphviz[3] module for Python. An architecture of code represented in the UML class diagram is shown in fig. 7.

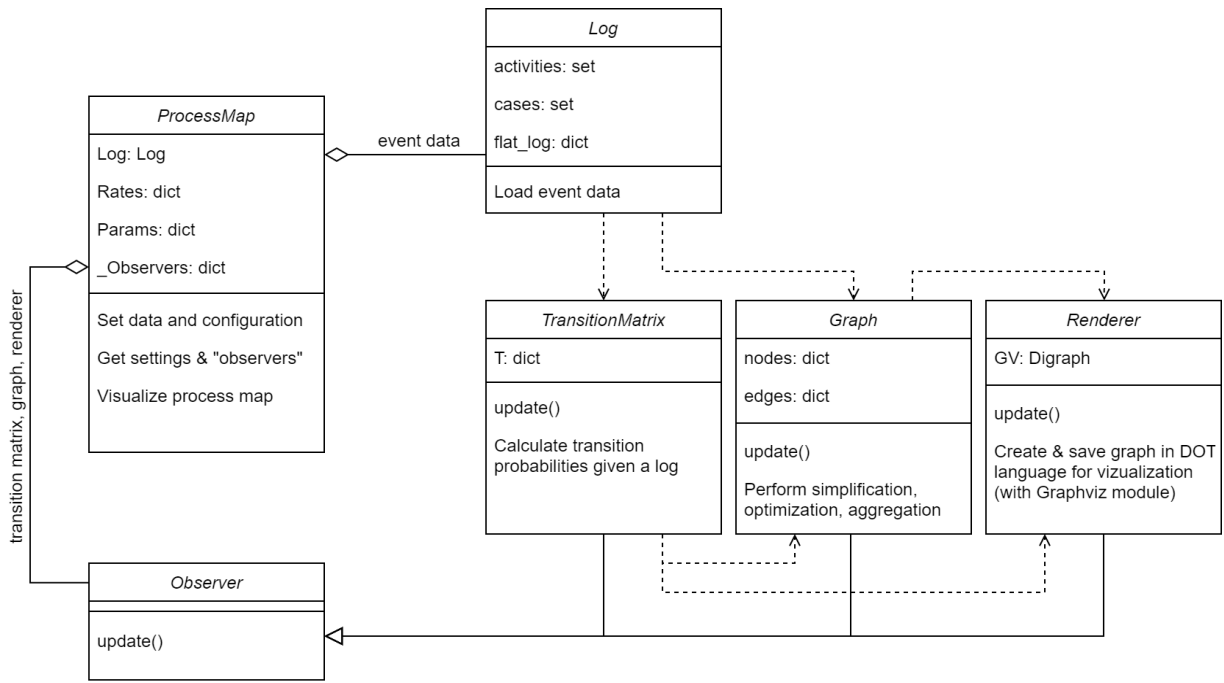

Figure 7 – UML class diagram for the main classes of ProFIT library

For a start working with a module, it is enough to pass a path to a directory with a log file as input in the *set_log* method. The module will produce an optimal model with default parameters. One can also tune model details "by hand" via the *set_rates* method that changes the activity and transition rates as well as change parameters via the *set_params* method, e.g., enable aggregation or optimization. Exploring data stored in the "observers" is possible by calling corresponding get-methods and visualizing a process map – by calling the *render* method. You can find out more details and code examples in the project repository at Github.

---

[2] https://github.com/Siella/ProFIT/
[3] https://pypi.org/project/graphviz/

## 5. Experimental study

### 5.1. Datasets

We consider two cases of process model discovery in the presented study, where the proposed solution was applied and validated. The first process to discover is remote monitoring of patients suffering from arterial hypertension provided by PMT Online[4] (a company specialized in the development of medical information systems and telemedicine systems). The second process is the daily activities of medical personnel in Almazov center[5] in Saint Petersburg, one of the leading cardiological centers in Russia. These cases in healthcare are seemed to be much better for exploring complexity because, as it is known, healthcare processes are highly diverse and uncertain on multiple levels of implementation. We present a summary of all datasets in table 1 and give a detailed description below.

First, we applied the proposed discovery technique to the monitoring event log consisted of 35,611 events, 272 cases corresponding to different patients, and 18 types of activities performed by operators, physicians, and nurses during monitoring of patients with arterial hypertension all over Russia within a telemedical system developed by PMT Online. We combined activity labels with corresponding resources to additionally reveal role interactions. If the same activities performed by different workers are aggregated, it is similar to clustering events, which are highly correlated in context sense. The remote monitoring program for patients with hypertension is as follows: the patients measure their blood pressure in-home on a regular basis, and each record made by a toolkit is transferred to a server, where data is processed then. There are several clinical events for medical staff that measurements may trigger. The main are "Red zone" and "Yellow zone" that notify about exceeding critical (emergency instance) and target (urgent instance) levels of blood pressure, respectively. These events have to be processed by operators and doctors, which may take some actions according to a scenario, e.g., contacting a patient by appointment or instantly. Usually, the "Red zone" events occur for patients that have not an appropriate treatment plan yet. When a health state normalized due to medications, "Yellow zone" appears rather than "Red zone", or it is possible a patient to be transferred to a therapy control program to maintain its blood pressure levels. There are also non-clinical events such as "New med. program" when a patient is registered for remote care, "Meas. missing" when data are not received by the server, etc. Ideally, when target levels are achieved, and the kit is returned to the monitoring provider (also are events), the program comes to an end with "Monitoring completed" event.

---

[4] pmtonline.ru (in Russian)
[5] almazovcentre.ru/?lang=en

A more challenging case study is discovering regular daily activities (workflow) of doctors and nurses from not process-aware hospital information system. Our colleagues from Almazov National Medical Research Centre provided us anonymized database with patient electronic health records covering COVID-19 treatment cases in their facility from March 2020 to June 2021. The dataset is a collection of fragmented medical records from patient history including patient id, event id, event description and associated record section name, timestamp, specialist name and type, department, record status, supplementary information as semi-structured text. We create an event log from raw data source following event log imperfection patterns [17]. They were form-based event capture, distorted label, collateral events, homonymous label, etc. From the obtained event log, we picked up one doctor and one nurse instances of process realizations. Therefore, we got two event logs where process case is defined by patient id.

Table 1 – Datasets summary

|  |  | **Monitoring Process** | **Nurse Workflow** | **Physician Workflow** |
|---|---|---|---|---|
| **Num. of cases** | | 272 | 165 | 43 |
| **Event classes** | | Clinical | Lab tests & Follow-up | Appointments |
| | | Non-clinical | Triage duties | COVID-19 treatment |
| **Num. of unique events** | | 18 | 19 | 29 |
| **Total num. of events** | | 35,611 | 1,042 | 1,077 |
| **Case length** | Max | 674 | 33 | 61 |
| | Min | 3 | 1 | 1 |
| | Mean | 131 | 6 | 25 |
| **Record duration** | | 355 days | 454 days | 377 days |

## 5.2. Complexity optimization

We aimed to investigate several measures of complexity and to make a comparison across the event logs and types of cycles folding. In this study, we considered four measures, such as an average degree, entropy, and two additional ones that we empirically derived in Section 4.2.

The measures (9) and (10) are structural. They indicate the relative size of the model, i.e., the ratio of the number of elements in the model and the number of possible or theoretical ones. The first is the number of edges presented in the model divided by the number of edges of a corresponding complete directed graph. This way, we can judge how close the graph structure to having all pairs of relations, which complicates its understanding. A complete graph does not imply loops, so, in this formula, we do not account for them in the number of edges even though they present in the model. The last measure (10) is the equally weighted sum of the activities and transitions ratios. "Start" and "end" events are not included in the set of nodes for this measure since they do not present in the log activities and are just auxiliary. Accordingly, in- and up-coming relations for the initial and terminal nodes are not included in the set of edges.

We plotted landscapes for each of the considered complexity measures. They visualize the complexity value and its relationships with activity and transition rates that are basic options in our algorithm to regulate the process model completeness. Process models in the area near the rate limits are mostly useless due to either very high complexity (with very high $r_a$ and $r_t$) or reduction of almost all significant activities in the model (with very low $r_a$ and $r_t$). We also revealed that there are no meta-states found for the event log of nurse workflow. So, only results for two event logs of the monitoring program and physician workflow are mainly discussed further. A summary of cycles found in the models is given in table 2. It should be pointed out that "start" and "end" events and associated with them relations are accounted for in the number of model elements. We also want to highlight that the maximum and the minimum numbers of cycles and meta-states are not always for the boundary levels, and the mean values are rounded down.

Table 2 – Cycles and meta-states found in the model

|  |  | Monitoring Process | Nurse Workflow | Physician Workflow |
|---|---|---|---|---|
| Num. of elements (activities/transitions) | Upper boundary (100/100) | 20/176 | 21/69 | 31/139 |
|  | Lower boundary (0/0) | 4/4 | 4/3 | 3/2 |
| Total num. of cycles | Max | 498 | 3 | 107 |
|  | Min | 1 | 0 | 0 |
|  | Mean | 102 | 0 | 18 |
| Num. of significant cycles | Max | 10 | 0 | 1 |
|  | Min | 1 | 0 | 0 |
|  | Mean | 7 | 0 | 1 |

We further provide and investigate fitness and complexity landscapes across the considered measures and event logs. At first sight, they may seem monotonic. However, intricate patterns in behaviors of real processes involve irregularities in the landscapes that are amplified with the presence of meta-states in the model. Nevertheless, we can still address the optimization problem within such conditions. The results of process model optimization are shown in figure 8, where a red marker is plotted to indicate optimal rates. One can see that all complexity measures can facilitate decreasing the transition rate, which directly affects the ability to comprehend a model effortlessly. Meanwhile, $H$ and $K_n$ allowed the maximum activity rate to be optimal in all cases. It is not appropriate if there are too many activities and highly varying process behaviors.

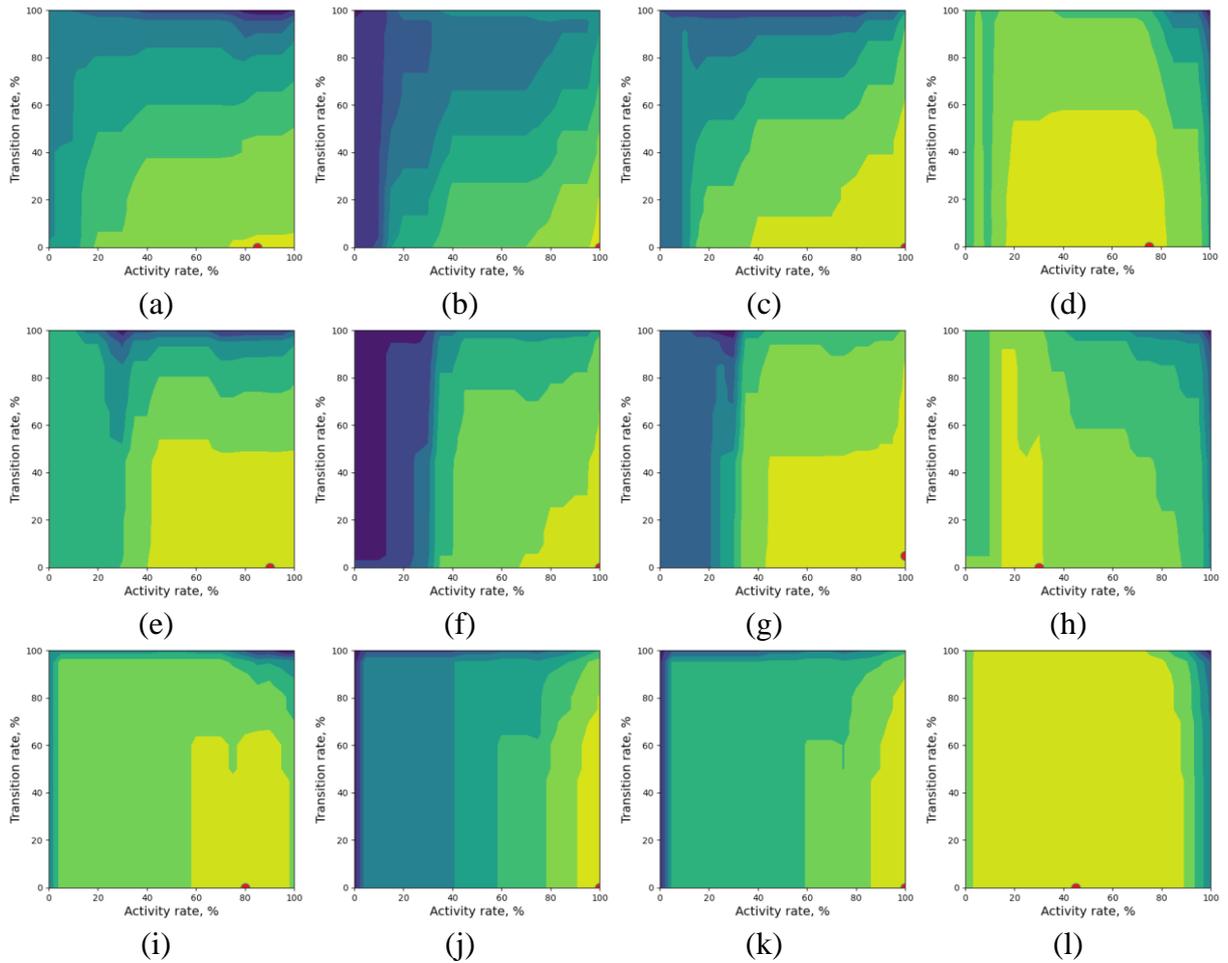

Figure 8 – Contour plots of target function (4) for the monitoring (a-d), physician (e-f), and nurse (i-l) event logs with complexity defined as $AD$ (a, e, i), $H$ (b, f, j), $K_n$ (c, g, k), and $R$ (d, h, l); $\lambda = 0.6$.

The complexity landscape of $R$ has a stepped form (fig. 9) due to a filtration principle: more rate values, more model elements. When we performed "outer" aggregation, the model complexity increased, because extra nodes as meta-states were added. Other types of aggregation hide all stand-alone events which compose significant cycles. That is why we observe lower complexity, especially where the maximum number of significant cycles is obtained. This applies in all proposed measures generally. Other complexity landscapes (fig. 10-12) are rather not step-like, and the measures depend mostly on the transition rate. $AD$ and $K_n$ are directly related to the number of edges and inversely related to the number of nodes, that makes it possible to lengthen paths remaining them simple to track. This is what we aimed to achieve: not to "cut off" the model and make it better to understand. As mentioned above, the average degree of a directed graph is just the ratio of the number of edges and the number of nodes. It seems that $AD$ and $K_n$ should have similar landscapes but with the first not being normalized and the second being penalized a lot for more activities in the model. However, there are some similarities in the forms of $K_n$ and entropy landscapes, which is an interesting observation. Indeed, the number of occurrences of the outcome 1 in the adjacency matrix is equal to the number of edges in the graph, and its probability is the number of edges divided by the number of nodes squared, almost as for $K_n$. However, there may be processes with all possible relationships of events. In this case, a complete process model will have an entropy of 0 and a complete graph ratio of 1.

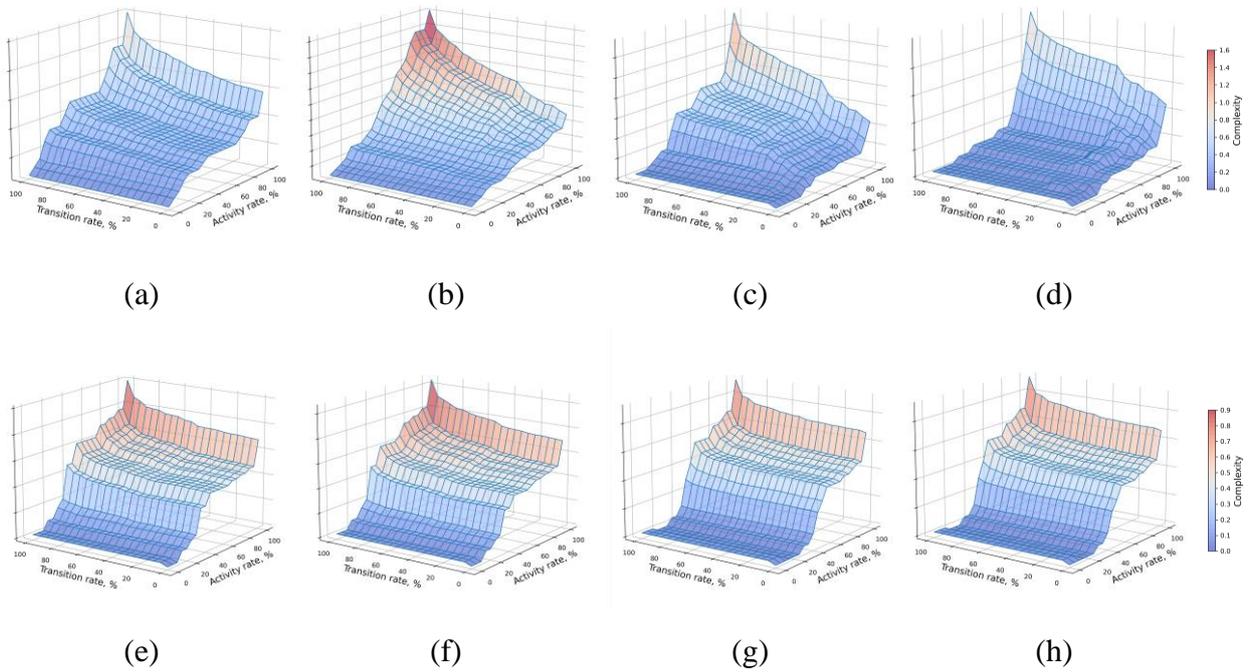

Figure 9 – Complexity landscapes of $R$ for the models of the monitoring program (a-d) and physician workflow (e-h) with no aggregation (a, e), outer joining (b, f), inner joining with $r_{all}$ (c, g), inner joining with $r_{freq}$ (d, h)

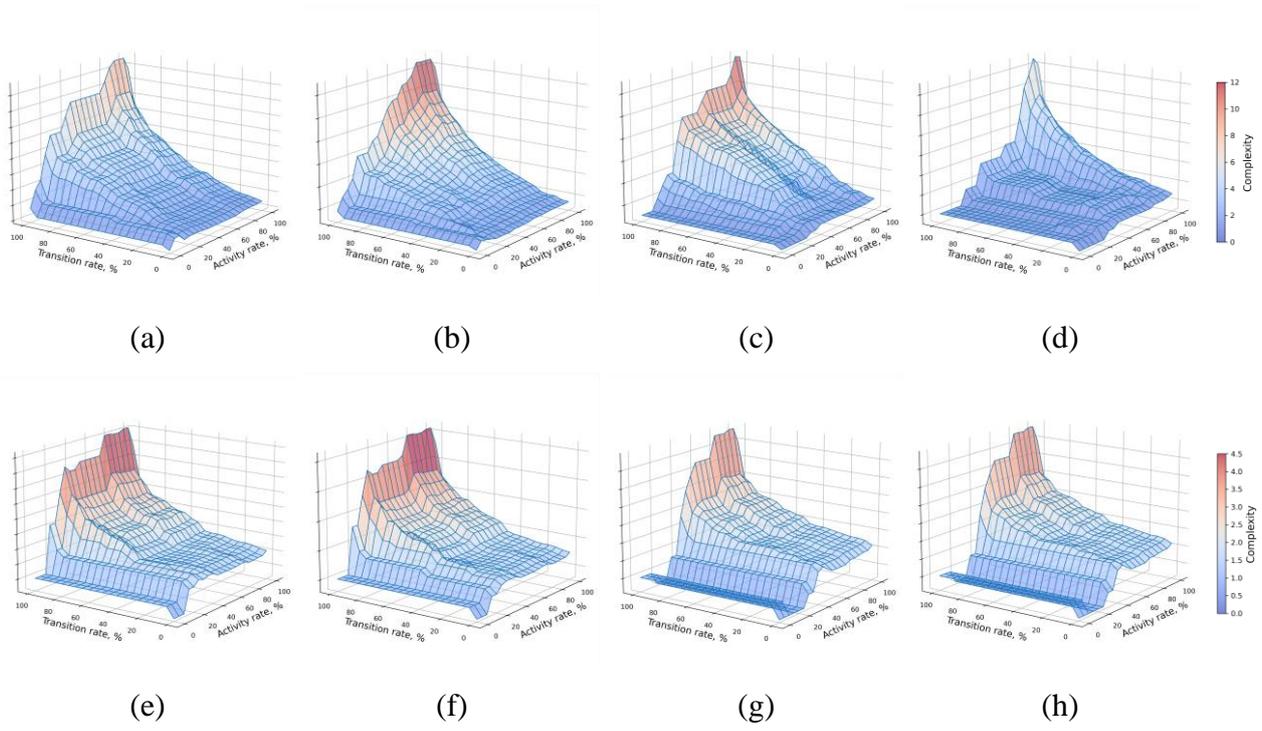

Figure 10 – Complexity landscapes of *AD* for the models of the monitoring program (a-d) and physician workflow (e-h) with no aggregation (a, e), outer joining (b, f), inner joining with $r_{all}$ (c, g), inner joining with $r_{freq}$ (d, h)

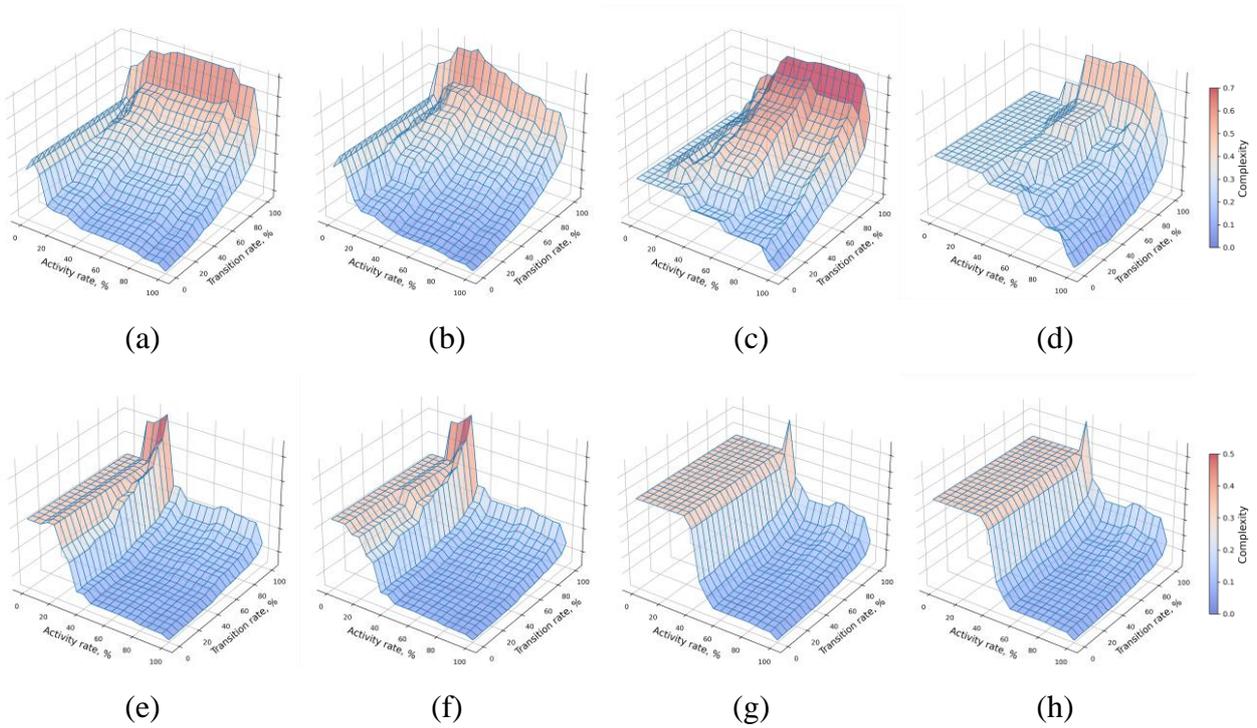

Figure 11 – Complexity landscapes of $K_n$ for the models of the monitoring program (a-d) and physician workflow (e-h) with no aggregation (a, e), outer joining (b, f), inner joining with $r_{all}$ (c, g), inner joining with $r_{freq}$ (d, h)

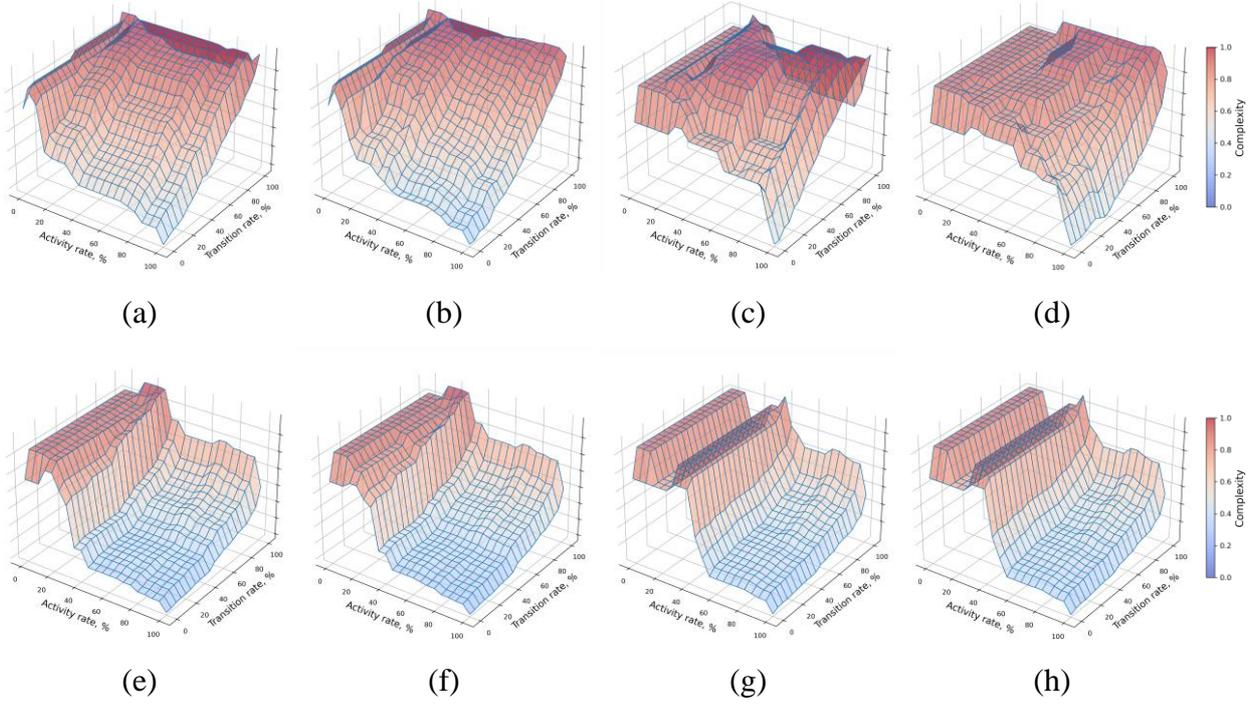

Figure 12 – Complexity landscapes of $H$ for the models of the monitoring program (a-d) and physician workflow (e-h) with no aggregation (a, e), outer joining (b, f), inner joining with $r_{all}$ (c, g), inner joining with $r_{freq}$ (d, h)

Table 3 – Process models optimization

| | Monitoring Process | Nurse Workflow | Physician Workflow |
|---|---|---|---|
| **No optimization** | | | |
| **Optimization, no aggregation** | | | |

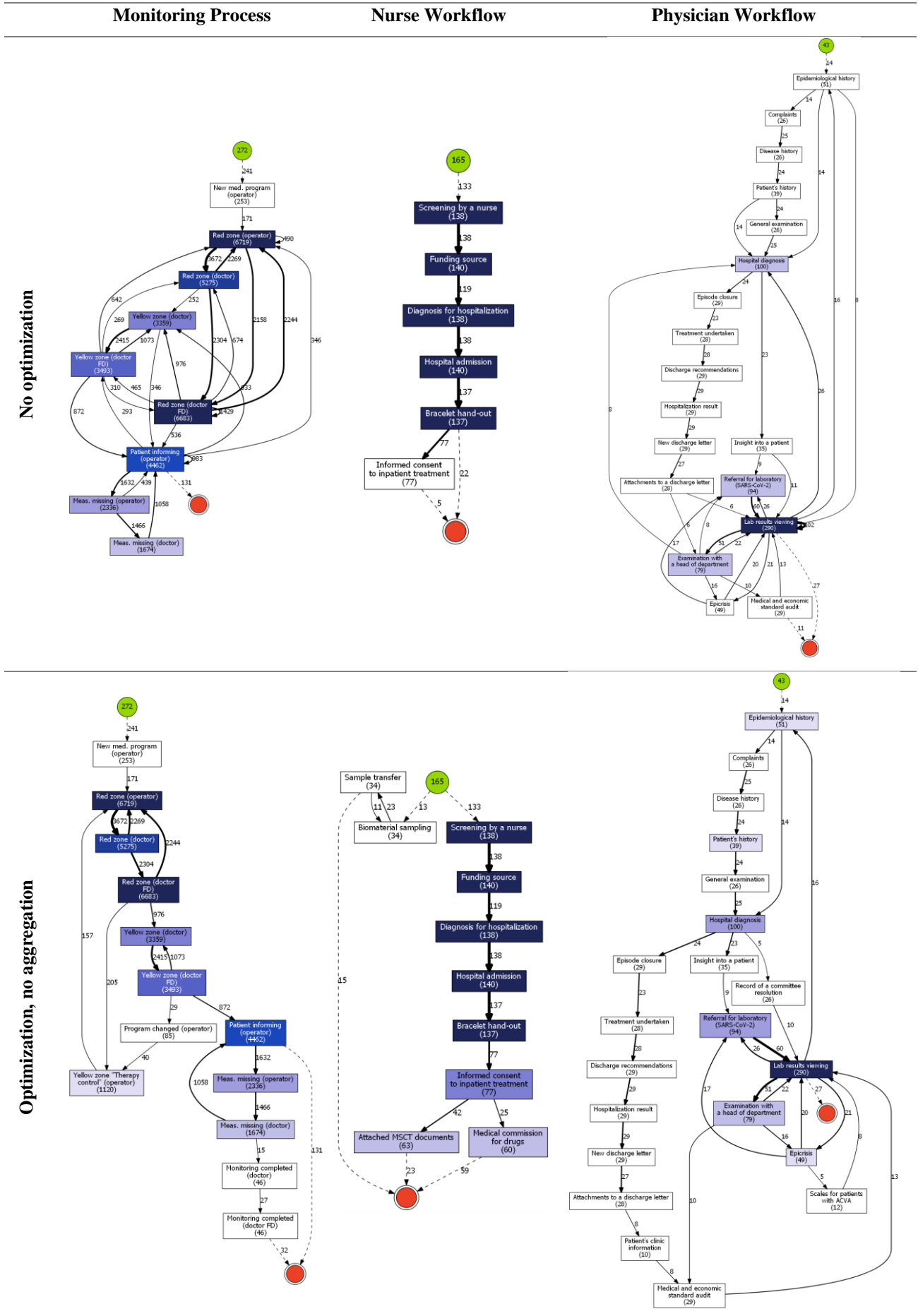

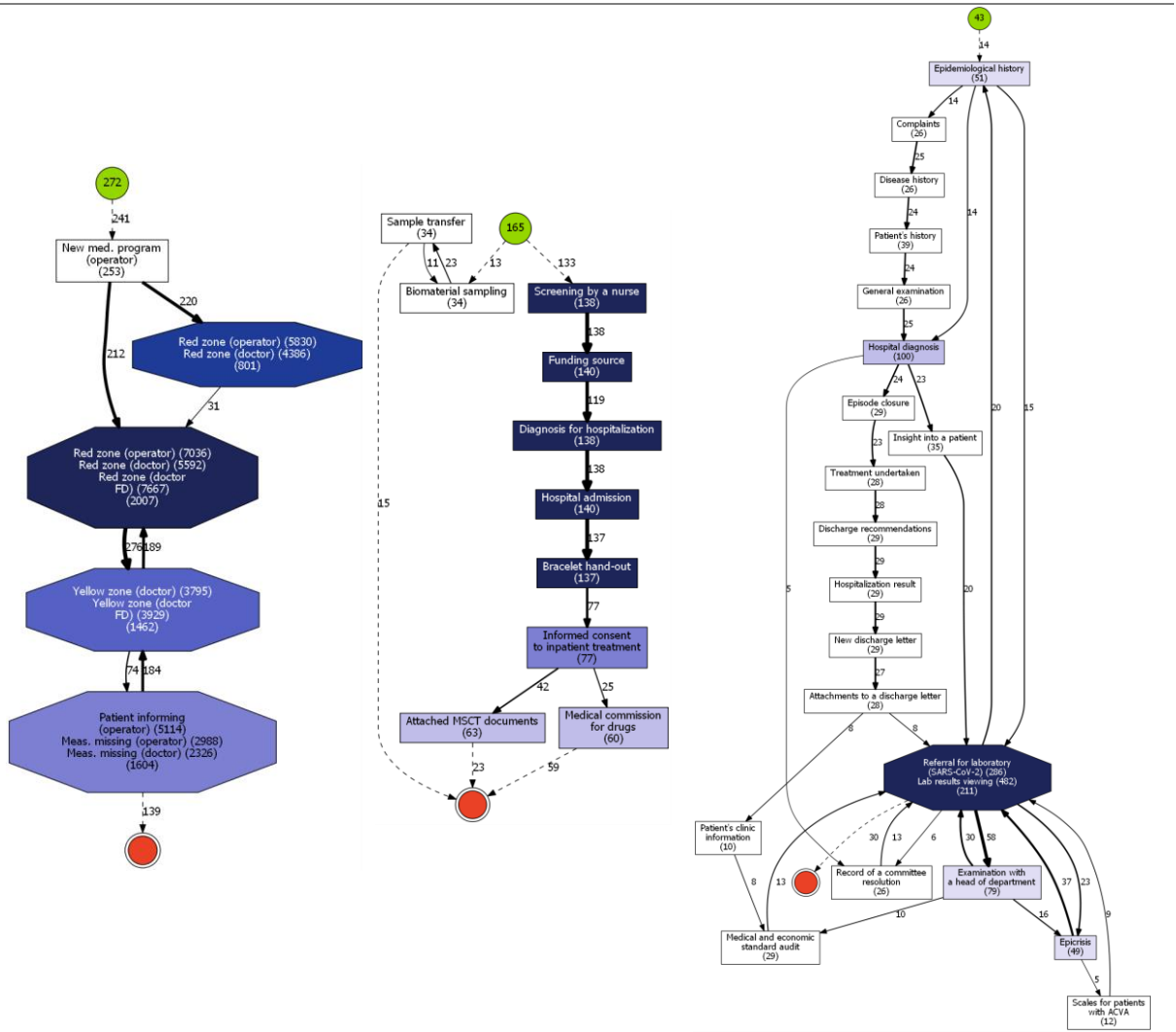

In the remainder of this section, we give the results of process models optimization with $C_R$ in table 3 and a summary across all proposed complexity measures in table 4. We consider 50/50 models as a baseline to compare them with optimal ones. These models are neither good nor bad, so we aim to see whether performing optimization and aggregation may have the odds in better results. We do not categorize process models on right and wrong but rather compare them on appropriateness for describing and comprehending the process. In our understanding, large process models, as well as too small, are not appropriate. We have mentioned the model size influences on the ability to understand it simple enough. Large process models cause cognitive difficulties for both analysts and common users, but meanwhile, a model with short paths of process execution may not reflect the complete process behaviors. In this regard, we succeed in discovering meaningful and intuitive process models (table 3). The optimization results using other complexity measures are shown in table 4, and some examples of the corresponding process models mined are in the appendix.

Table 4 – Optimized process models summary

| | | Monitoring Process | | | | Nurse Workflow | | | | Physician Workflow | | | |
|---|---|---|---|---|---|---|---|---|---|---|---|---|---|
| **Agg.** | | NA | O | I,$r_{all}$ | I,$r_{freq}$ | NA | O | I,$r_{all}$ | I,$r_{freq}$ | NA | O | I,$r_{all}$ | I,$r_{freq}$ |
| **50/50** | $r_a$ | 50 | 50 | 50 | 50 | 50 | 50 | 50 | 50 | 50 | 50 | 50 | 50 |
| | $r_t$ | 50 | 50 | 50 | 50 | 50 | 50 | 50 | 50 | 50 | 50 | 50 | 50 |
| | F | 0.91 | 0.91 | 0 | 0 | 0.64 | 0.64 | 0.64 | 0.64 | 0.93 | 0.93 | 0.40 | 0.40 |
| | AD | 2.73 | 3.13 | 5.09 | 1.60 | 1.00 | 1.00 | 1.00 | 1.00 | 1.85 | 1.90 | 1.80 | 1.80 |
| | H | 0.80 | 0.74 | 1.00 | 0.90 | 0.54 | 0.54 | 0.54 | 0.54 | 0.44 | 0.45 | 0.53 | 0.53 |
| | $K_n$ | 0.25 | 0.20 | 0.48 | 0.35 | 0.14 | 0.14 | 0.14 | 0.14 | 0.09 | 0.10 | 0.12 | 0.12 |
| | R | 0.34 | 0.50 | 0.40 | 0.10 | 0.21 | 0.21 | 0.21 | 0.21 | 0.45 | 0.45 | 0.32 | 0.32 |
| **Optimized** AD | $r_a$ | 85 | 85 | 85 | 85 | 80 | 80 | 80 | 80 | 90 | 90 | 90 | 90 |
| | $r_t$ | 0 | 0 | 0 | 0 | 0 | 0 | 0 | 0 | 0 | 0 | 0 | 0 |
| | F | 0.96 | 0.97 | 0.02 | 0.02 | 0.85 | 0.85 | 0.85 | 0.85 | 0.97 | 0.97 | 0.54 | 0.54 |
| | J | 1.40 | 1.53 | 1.29 | 1.29 | 1.17 | 1.17 | 1.17 | 1.17 | 1.43 | 1.42 | 1.50 | 1.50 |
| H | $r_a$ | 100 | 100 | 100 | 100 | 100 | 100 | 100 | 100 | 100 | 100 | 100 | 100 |
| | $r_t$ | 0 | 0 | 0 | 0 | 0 | 0 | 0 | 0 | 0 | 0 | 0 | 0 |
| | F | 0.96 | 0.97 | 0.07 | 0.07 | 0.99 | 0.99 | 0.99 | 0.99 | 0.98 | 0.98 | 0.54 | 0.54 |
| | J | 0.38 | 0.34 | 0.44 | 0.44 | 0.40 | 0.40 | 0.40 | 0.40 | 0.29 | 0.28 | 0.30 | 0.30 |
| $K_n$ | $r_a$ | 100 | 100 | 100 | 100 | 100 | 100 | 100 | 100 | 100 | 100 | 100 | 100 |
| | $r_t$ | 0 | 0 | 0 | 0 | 0 | 0 | 0 | 0 | 5 | 5 | 5 | 5 |
| | F | 0.96 | 0.97 | 0.07 | 0.07 | 0.99 | 0.99 | 0.99 | 0.99 | 0.98 | 0.98 | 0.55 | 0.55 |
| | J | 0.08 | 0.07 | 0.10 | 0.10 | 0.08 | 0.08 | 0.08 | 0.08 | 0.05 | 0.05 | 0.06 | 0.06 |
| R | $r_a$ | 75 | 75 | 75 | 75 | 45 | 45 | 45 | 45 | 30 | 30 | 30 | 30 |
| | $r_t$ | 0 | 0 | 0 | 0 | 0 | 0 | 0 | 0 | 0 | 0 | 0 | 0 |
| | F | 0.94 | 0.95 | 0.02 | 0.02 | 0.64 | 0.64 | 0.64 | 0.64 | 0.62 | 0.62 | 0.06 | 0.06 |
| | J | 0.32 | 0.46 | 0.16 | 0.13 | 0.21 | 0.21 | 0.21 | 0.21 | 0.14 | 0.14 | 0.04 | 0.04 |

We recall that the aggregation step follows the optimization in the algorithm's workflow (fig. 2), i.e., models with meta-states are obtained within fixed optimal parameters ($r_a$ and $r_t$) found before. We additionally defined how fitness is calculated for the aggregated models: if $(u, v) \in E$, $u \in V^-$, $v \in \tilde{V}$, then we add $(u, v_i)$, $\forall v_i$ in $v$, and all transitions composing $v$ to the list of edges "presenting" in the model. Other cases ($(v, u)$ and $(v, v')$, $v' \in \tilde{V}$) are treated by analogy. It was done to demonstrate the quantified results for all models that may be discovered across different complexity measures, aggregation types, event data. This way, one can get a better insight into the comprehension difficulty and precise of the discovered models via numerical comparison in addition to a visual assessment.

## 6. Discussion

The experimental study shows that the proposed approach can be applied in various conditions and problem domains and diverse structure of process maps. The optimization procedure enables the construction of explicit and understandable process maps with good coverage of the event log. The optimized process maps for the considered cases reveal key sequences of events and demonstrate a good reflection of the processes' nature. One of the challenging issues discovered within the study is managing cyclic meta-states as a part of optimization and interpretation procedures.

The revealed complexity landscapes (see fig. 9-12) after the introduction of aggregation procedures become highly "rugged". That leads to the appearance of multiple local optimums. Although, these optimums may be considered as stable and interpretable process maps, they appear to domain experts in terms of covering or not-covering certain field-specific states. Within our approach, we were focused on reaching higher interpretability of process maps from that particular point of view. Thus, to our beliefs, this issue goes beyond the problem of global optimization. The presence or absence of particular meta-states should be considered from the domain-specific point of view with a further interpretation, which can impose restrictions on a global optimization problem during the process discovery within the proposed approach.

The number of meta-states varies over the parameter space significantly (see, e.g., fig. 13(a) for the monitoring process). To analyze the structure of meta-states discovered in different areas of the parametric space, we have identified the meta-states appearance for the monitoring program case. Within a basic grid search, we have discovered 15 possible combinations of meta-states (see fig. 13(b) and fig. 13(d) for a description of states). Still, the more important issues can be revealed when considering a structure of adjacent areas. In fig. 13(c), we introduce a graph structure showing the transitions between combinations by adding meta-states (e.g., edge "+CF|HD" means that the moving from combination $C_{14}$ to combination $C_{12}$ is reflected in adding two cycles "CF" and "HD" – see fig. 13(d) caption for the interpretation). Several measures can be introduced to identify the relevance of the combination to the actual process, e.g., by assessing coverage area in parameter space or centrality measure in the proposed graph structure. Here, one can select combinations $C_2$, $C_9$, $C_{14}$ by its centrality and high coverage of parameter space. Moreover, $C_2$, $C_9$ have implicit evidence of its consistency as transitions from them loose multiple meta-states with increasing complexity (see, e.g., transitions to combinations $C_3, C_4, C_5, C_6, C_8$). Thus, in this case, the combinations can be selected to define the area for the optimization within the parametric space.

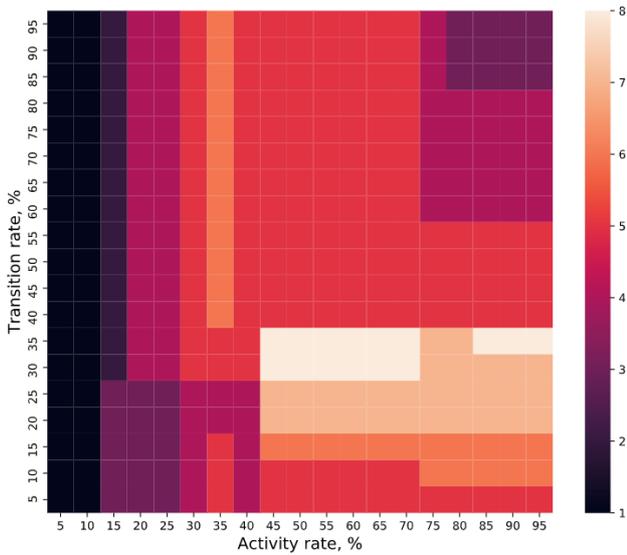
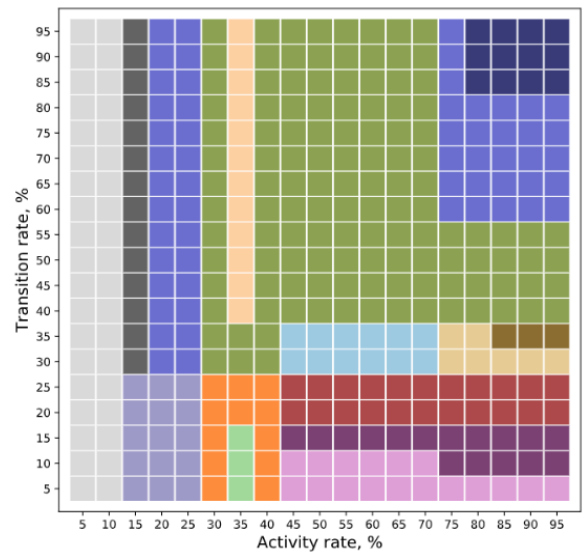

(a)  (b)

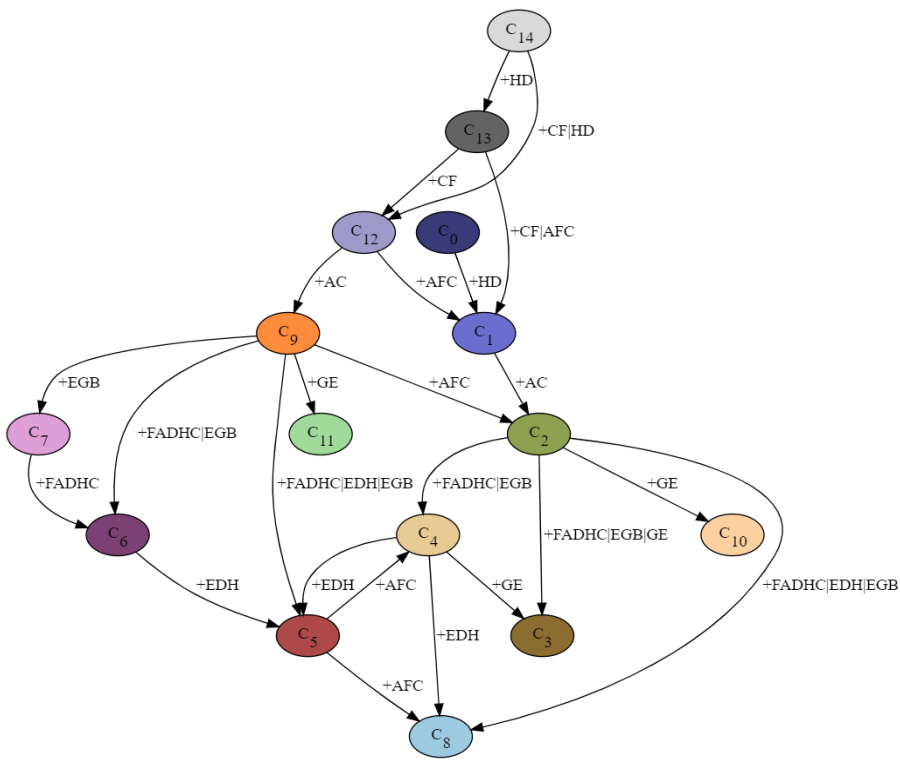
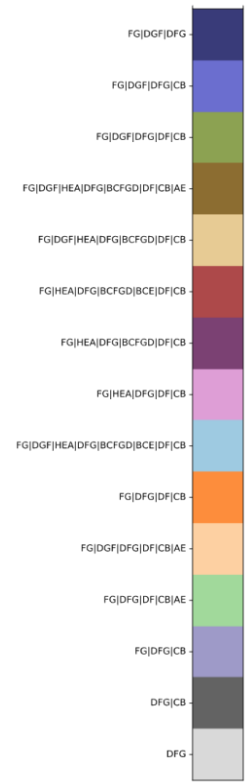

(c)  (d)

Figure 13 – Meta-state combinations (a) number of significant cycles; (b) covered area; (c) transition states; (d) legend (Meas. missing (operator): A, Yellow zone (doctor): B, Yellow zone (doctor FD): C, Red zone (doctor FD): D, Patient informing (operator): E, Red zone (operator): F, Red zone (doctor): G, Meas. missing (doctor): H)

We treat the discovered issues as insights for the development of the proposed approach towards higher domain-specific interpretability and consistency of process models discovered automatically. Along with the other interpretability issues, e.g., tuning process map layout for better human comprehension, we consider them as a direction for further work.

## 7. Conclusion and future works

In this paper, we presented the algorithm for automatic process model discovery and the method of process model abstraction and interpretation. We defined the problem of process model optimization to achieve the balance between two terms: model correctness for event data, i.e., fitness, and model complexity, i.e., a measure of its comprehension difficulty. We proposed several complexity measures in the experimental part of the study and conducted a comprehensive analysis of their influences on the model look and its parameters. We demonstrated our solution validity on the event logs from the healthcare domain. Still, the algorithm is general-purpose and is adaptable to different fields and tasks.

In future studies, we plan to continue the work on the project and extend its functionality. One of the promising directions for the development is extending interpretability capabilities within the solution with different knowledge sources, including formal knowledge and data mining. Machine learning models or Hidden Markov models, e.g., can be used to interpret meta-states found in the process models or, vice versa, knowledge mined from the event logs can be employed in predictive modeling. We are also interested in the integration and application of the developed solution in various problem domains. We have a belief of much room in a process mining application, that can lead to interesting and valuable results.

**Acknowledgments.** This work was supported by the Ministry of Science and Higher Education of Russian Federation, goszadanie no. 2019-1339.